\documentclass[letterpaper, 10 pt, conference]{ieeeconf}  

\IEEEoverridecommandlockouts                              

\usepackage{graphics} 
\usepackage{epsfig} 
\usepackage{mathptmx} 
\usepackage{amsmath} 
\usepackage{amssymb}  
\usepackage{algorithm}
\usepackage{algorithmic}
\usepackage{array}
\usepackage{booktabs}

\title{\LARGE \bf
Retractable Prosthesis for Transfemoral Amputees Using Series Elastic Actuators and Force Control
}

\author{Elena Galbally$^{1}$, Frank Small$^{1}$, and Ivan Zanco$^{1}$
\thanks{$^{1}$Mechanical Engineering Department, University of Maryland}%
}

\begin{document}

\maketitle
\thispagestyle{empty}
\pagestyle{empty}

\begin{abstract}
We present a highly functional and cost-effective prosthesis for transfemoral amputees that uses series elastic actuators. These actuators allow for accurate force control, low impedance and large dynamic range. The design involves one active joint at the knee and a passive joint at the ankle. Additionally, the socket was designed using mirroring of compliances to ensure maximum comfort. 
\end{abstract}

\section{Introduction}
Transfemoral amputation involves the loss of two joints: knee and ankle. Both of these joints play crucial roles in enabling efficient human motion. The former is a rotational joint that permits flexion and extension of the leg as well as a slight internal and external rotation [1]. The later generates most of the power needed to walk and is critical for shock absorption and balance. 

In order to allow smooth movements, avoid gait asymmetries, and prevent falls, a transfemoral prosthesis must be able to address the following needs:
\begin{enumerate}
  \item Provide enough power for the foot to safely clear the ground.  
  \item Adapt to the user's walking pattern and speed.
  \item Function in different terrains and environmental conditions. 
  \item Provide robust footing and adequate shock absorption.
  \item Comfort.
\end{enumerate}

For 50\% of transfemoral prosthetic users, gait is not automatic. In other words, they must think every step they make \cite{GauthierGagnon99UseOfTransfemoralProsth}. Nowadays, most high$-$tech prosthesis are too expensive for the average user and commercial designs are not functional enough to allow complex locomotor activities.Thanks to our material, sensor, and actuation selection, the design presented in this paper provides an optimal balance between cost and functionality. Hence, this prosthesis for transfemoral amputees could potentially be commercialized in low-income countries where high-tech bionic devices are not an option. 

\begin{figure}
\begin{center}
\includegraphics[width=60mm]{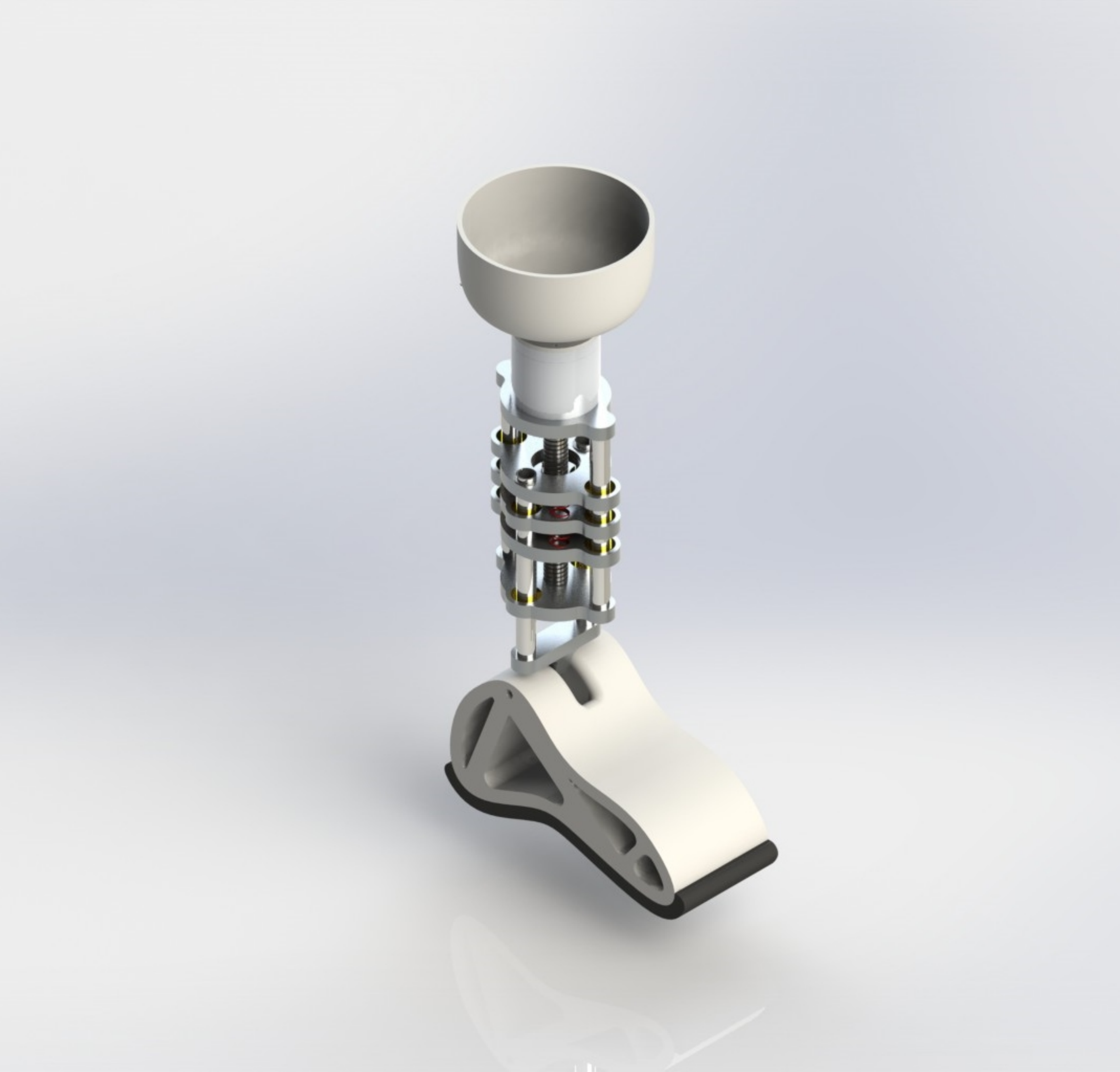}
\end{center}
\caption{Transfemoral prosthesis prototype: uses a brushless DC motor rigidly connected to a ballscrew which drives the linear motion. The socket presented in the figure is a simplified representation of our final design. See section \ref{SocketDesign}.}
\label{fig_prosthesis}
\end{figure}

The prosthesis that are currently being used in sub-developed countries are passive and uncomfortable \cite{Pearlman08Cheap}. Therefore, even though our design involves only one active joint and a basic sensory-actuator system, it will undoubtedly be a major improvement with regard to existing devices of a similar price.The prototype is shown in Fig. \ref{fig_prosthesis}. 


\section{Related Work}

\subsection{Actuators}
There has been considerable progress in developing actuators for robotics in the past few decades. In particular, Series Elastic Actuators were developed by Pratt and Williamson \cite{Pratt95SeriesElasticAct, Matthew95SerElastActThesis} who directly measured the strain of a spring in series with transmission and actuator output. Since the spring deforms a signicant amount, the fidelity compared to typical strain gauge structures for force control is much higher. However, their motion bandwith is rather small, but biomimetic actuators can trade off small motion bandwidth for good force control. 

The highest performance force controlled actuator has been a brushless DC motor rigidly connected to a robot link, also know as direct-drive \cite{Asada87DirectDriveBOOK}. These actuators eliminate friction and backlash, typical of motors with transmissions. To compensate for the loss of transmission, direct-drive actuators must be large in order to achieve adequate torque. This means increased motor mass and cost.

Although, Series Elastic Actuators may not have as large a dynamic range as comparable direct drive actuators, the weight savings are well worth the trade off for our intended weight sensitive application.

\subsection{Robots}
Several biomimetic robots using Series Elastic Actuators have been constructed and demonstrated. For instance, Spring Turkey \cite{pratt97SpringTurkey} and SpringFlamingo \cite{pratt98SpringFlamingo} developed by MIT use linear drive Series Elastic Actuators.

Other laboratories and companies have also worked on walking robots. One of the most remarkable and well known being Boston Dynamic's Big Dog. A quadruped robot desingned to achieve animal$-$like mobility on rough and rugged terrain, terrain too difficult for any existing vehicle. It uses low-friction
hydraulic cylinders regulated by two-stage aerospace-quality
servovalves for actuation. Each actuator has sensors for joint position and
force \cite{BD08BigDog}.

\subsection{Force Control}
In
highly unstructured environments, force controlled robots
that can comply to the surroundings are desirable \cite{Arumugom09ModelingLinearActutator}.

An ideal force‐controllable actuator would be a
perfect force source. In a perfect force
source, impedance is zero (completely back drivable),
stiction is zero, and bandwidth is infinite.

By adding Series Elasticity to these
conventional systems, a force‐controllable actuator
with low impedance, low friction, and good
bandwidth will result.

Thanks to their ability to closely approximate a pure force source, Series Elastic Actuators
lead to a much more accurate force control than other traditional technologies such as \cite{Arumugom09ModelingLinearActutator}:
\begin{itemize}
\item Direct drive actuation.
\item Current control with a geared actuator.
\item Current control with low$-$friction cable drive transmissions.
\item Load cells with force feedback.
\item Fluid pressure control.
\end{itemize}

\subsection{Socket}

Scientific studies have been conducted to quantify
attributes that may be important in the creation of more functional
and comfortable lower-limb prostheses. The prosthesis
socket, a human-machine interface, has to be designed properly
to achieve satisfactory load transmission, stability, and efficient
control for mobility. The biomechanical understanding of
the interaction between prosthetic socket and the residual limb
is fundamental to such goals \cite{Arthur01SocketReview}.

Some
early designs of the prosthetic socket, such as the “plugfit,”
took the form of a simple cone shape, with very little rationale for the design.
With an understanding of the residual limb anatomy and
the biomechanical principles involved, more reasonable
socket designs, such as the patellar tendon bearing (PTB)
transtibial socket, and the quadrilateral transfemoral suction
socket were developed following World War II \cite{Radcliff61Patellar_Socket, Radcliff55Fitting_Socket}.

By the 1980s, the so-called hydrostatic
weight-bearing principle and the total surface bearing
(TSB) concept were introduced. Examples include the
silicone suction socket \cite{Fillauer89Socket} and ICEROSS \cite{Kristinsson93Socket}, as well as
those incorporating the use of interfacing gel-like
materials.

The most radical of new prosthetic developments is
certainly direct skeletal attachment of limb prostheses
through osseointegrated implants. This method completely
obviates the need for the prosthetic socket through percutaneous
titanium fixtures that transfer load from the
prosthesis directly to the skeletal bone. While it may
seem that osseointegration renders moot any discussion
of prosthetic interfaces, even this radical advance in the
state of the art only changes the location and type of the
interface problem. New challenges arise from the
metal/bone and metal/skin interfaces \cite{Arthur01SocketReview}.

\section{Approach}
\begin{figure}
\begin{center}
\includegraphics[width=60mm]{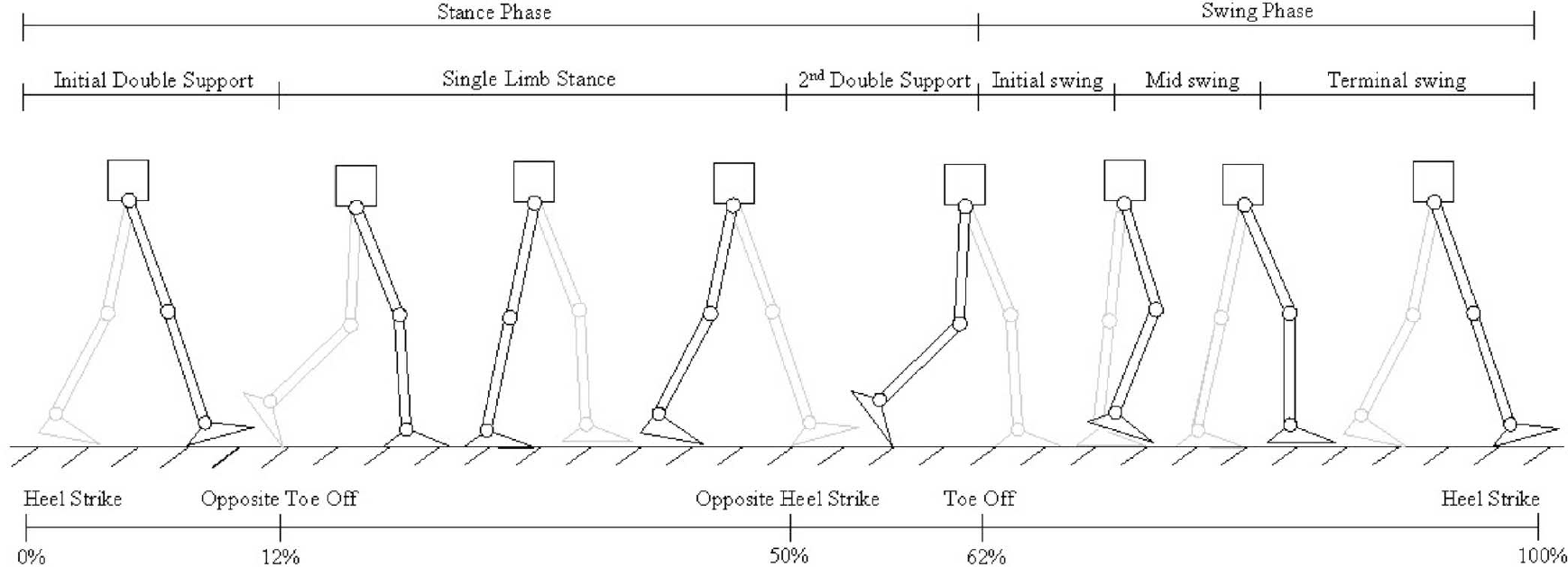}
\end{center}
\caption{Human walking gait through one cycle, beginning and ending at heel strike. Percentages showing contact events are given at their approximate location
in the cycle. Taken from \cite{Dollar008LowerExtremStateofArt}.}
\label{fig_gaitPaper}
\end{figure}

\begin{figure}
\begin{center}
\includegraphics[width=60mm]{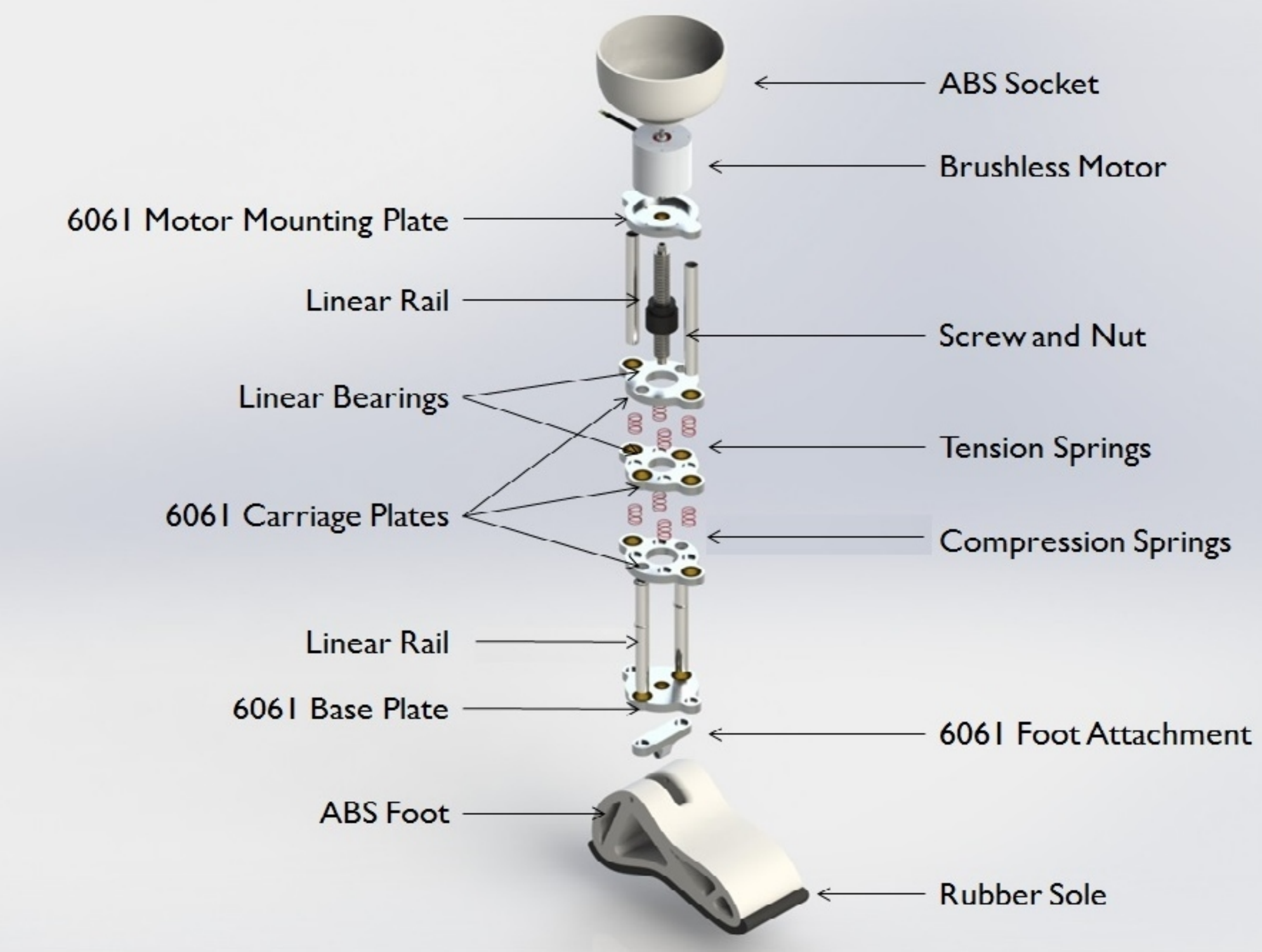}
\end{center}
\caption{Exploded view of the prototype showing all of the components.}
\label{fig_explosionNAmes}
\end{figure}

The design aims to enable the user to walk with ease at a low price. In order to do so, understanding the biomechanics of human walking is crucial, see Fig. \ref{fig_gaitPaper}. Our main innovation is using linear actuation instead of rotatory for the knee joint. This significantly lowers the price and mechanical complexity while maintaining high functionality. One of the few downsides of Series Elastic Actuators is their relatively small range of motion. However, we can see in Fig. \ref{fig_gaitPaper} that we usually do not lift the foot off the ground more than a couple centimeters in each stride. Additionally, thanks to the retractability feature, this prosthesis can be easily adapted to users with a wide range of heights. In other words, it is a "one fits all" design and, therefore, extremelly cost$-$efficient. The exploded view in Fig. \ref{fig_explosionNAmes} helps get a better grasp of how the mechanism works and the different parts that it comprises.


\section{Series Elastic Actuators}
Series Elastic Actuators provide many benefits in force control of robots in unconstrained environments. These benefits
include high force fidelity, extremely low impedance, low friction, and good force control bandwidth. Series Elastic Actuators employ
a novel mechanical design architecture which goes against the common machine design principal of “stiffer is better”. A compliant
element is placed between the gear train and driven load to intentionally reduce the stiffness of the actuator. A position sensor
measures the deflection, and the force output is accurately calculated using Hooke’s Law (F=Kx). A control loop then servos the
actuator to the desired output force. The resulting actuator has inherent shock tolerance, high force fidelity and extremely low
impedance \cite{Arumugom09ModelingLinearActutator}. Fig. \ref{fig_actuatorSchematics} shows the general architecture of Series Elastic Actuators. 

Note that Series Elastic Actuators are topologically similar to any motion actuator with a load sensor and closed loop control system. In fact, the main components of this actuation system are:
\begin{enumerate}
\item Motor
\item Transmission
\item Spring
\item Sensor
\item Controller
\end{enumerate}

Our actuator design can be seen in Fig. \ref{fig_carriage}. In the following sections we will study different design parameters that must be taken into account when choosing each of the components.


\begin{figure}
\begin{center}
\includegraphics[width=60mm]{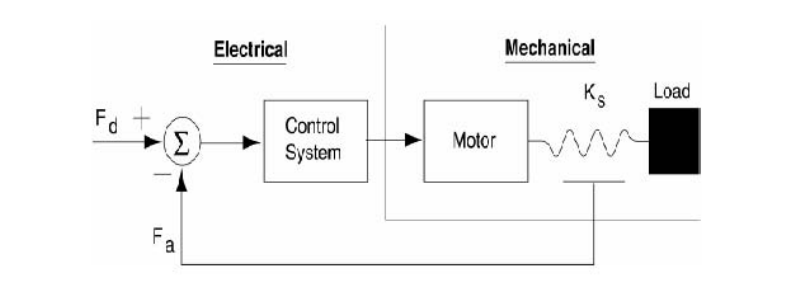}
\end{center}
\caption{Schematic diagram of a Series Elastic
Actuator. A spring is placed between the motor and
the load. A control system servos the motor to reduce
the difference between the desired force and the
measured force signal. Taken from \cite{Arumugom09ModelingLinearActutator}.}
\label{fig_actuatorSchematics}
\end{figure}


\begin{figure}
\begin{center}
\includegraphics[width=60mm]{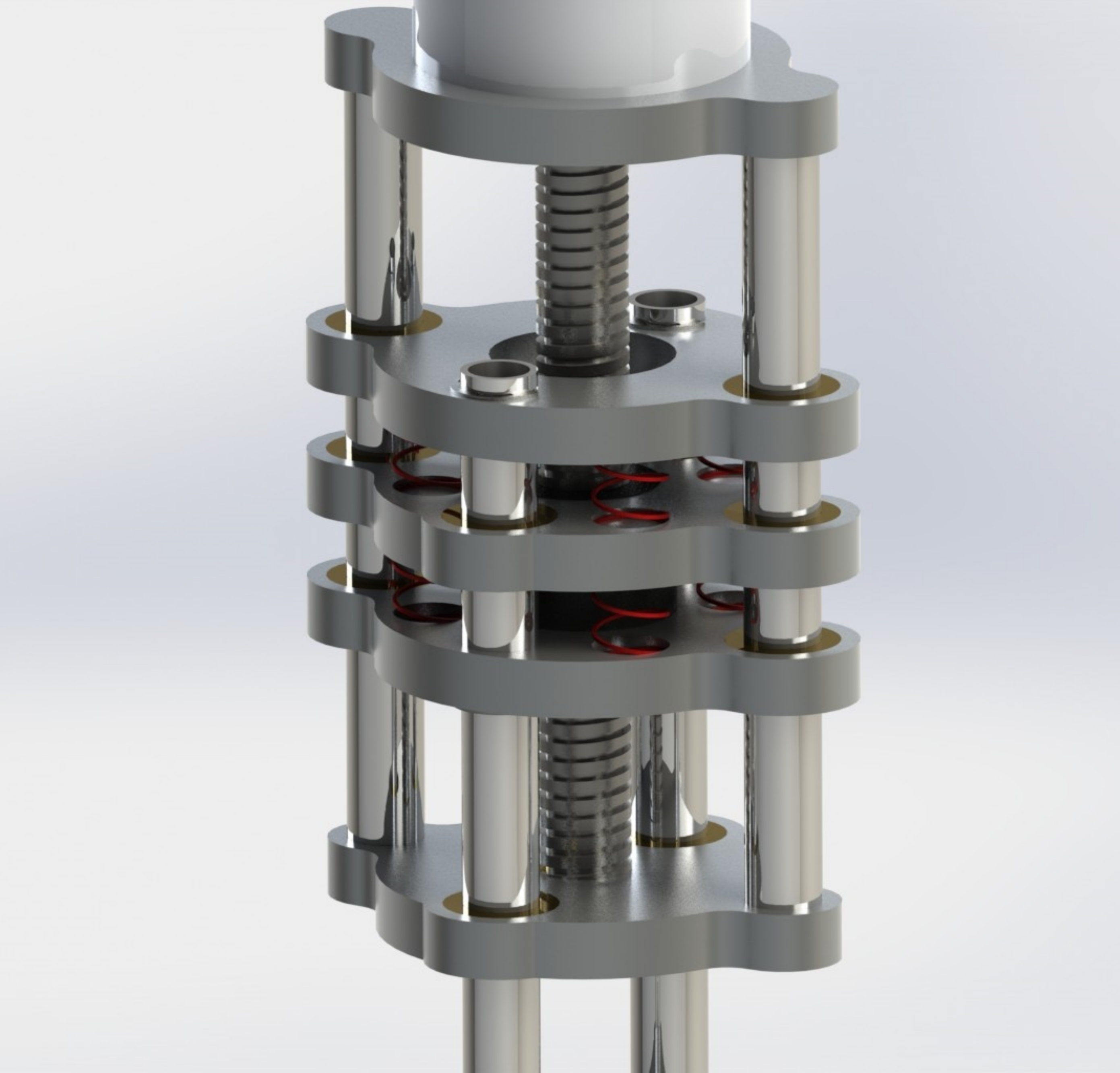}
\end{center}
\caption{CAD rendering of the Series Elastic Actuator used in the knee joint.}
\label{fig_actuatorSchematics}
\end{figure}


\subsection{Motor Selection} 

When selecting a servomotor there are many parameters we must take into account. It is worth highlighting the following:

\begin{itemize}
\item Rated Voltage: the voltage it is most efficient while running.
\item Operating current: average amount of current the motor is expected to draw under a typical torque.
\item Average Power = Rated Voltage x Operating Current
\item Stall current: maximum amount of current the motor will ever draw, and hence the maximum amount of power. It is measured by powering up the motor and then applying enough torque to force it to stop rotating. 
\item Stall torque: torque required to stop the motor from rotating.
\item Operating torque: the torque the motor was designed to provide. Usually it is the listed torque value (it is typically intended to be applied at 1 cm from the shaft).
\item Power requirements = Load Torque x Speed
\item Operating speed: refers to the maximum speed at which motor can turn: measured in seconds per 60º. If it is 0.5 seconds per 60º, then it will take 1.5 seconds to turn 180º.
\end{itemize}



 	\begin{table}[]
 		\begin{center}
 		\caption{Motor Specifications}
 		\label{table_motor}
 		\begin{tabular}{|c|c|}
 			\hline
 			Operating Speed & $\omega_m=4790$ rpm  \\
 			\hline
 			Operating Torque & $\tau_m=1.69$ N$\times$m \\
 			\hline
			Voltage Supply  & $V = 50$ V\\
 			\hline
 			Weight & $3.3$ kg\\
 			\hline
 		\end{tabular}
 		\end{center}
 	\end{table}


Since our mechanism uses a servomotor coupled to a ball$-$screw (see section \ref{ballScrew}) we used the following approach:

\begin{enumerate}
\item Calculate the torque (\ref{eqn_torqueBS}) and speed (\ref{eqn_speedBS}) requirements of the ball$-$screw:

\begin{eqnarray}
    \tau_{BS} &=& \frac{F \times L}{2 \times \Pi \times \eta} = 1.18  (N \times m) \label{eqn_torqueBS} \\
	\omega_{BS} &=&  \frac{V_L}{L} = 3600 (rpm) \label{eqn_speedBS}
\end{eqnarray}


Where: 
\begin{itemize}
\item F = Force = 300 lb (1334 N) $\rightarrow$ Assuming the person's weight is 200lb and that during walking each leg experiences 1.5 x body weight \cite{ForceWalking}.
\item L = Screw Lead = 5 mm/rev $\rightarrow$ Linear displacement of nut for one revolution of screw
\item $\eta$ = Efficiency = 0.9 \cite{BallScrewEfficiency}
\item $V_L$ = Linear Speed = 18000 mm/min

\end{itemize}

\item Select a motor whose operating speed and torque exceed the ball$-$screw requirements:
We chose the MOOG BN34 – 55 EU – 02LH fabricated by Moog Inc.The specifications can be seen in Table \ref{table_motor}.

We decided to use a brushless DC motor because it has been shown to minimize the motor friction seen through the transmission \cite{Robinson99LinearBiomimeticActuatorMIT}. 
Keeping friction and motor saturation low is always desirable. 

Additionally, we used a frameless motor
conguration where the motor magnets are mounted
directly onto an extended ballscrew shaft instead of
using a coupling, gears, or a belt drive. The main goal of this was to keep transmission dynamics at a minimum, as it is one of two limiting factors in using high
feedback gain \cite{eppinger87BandwidthLimit}.

The other limiting component to achieve high feedback
gain is the sensor. See section \ref{sec_sensor}.

\end{enumerate}

\subsection{Sensors: Linear Potentiometers} \label{sec_sensor}
The sensor needs to directly
measure the spring deflection. This insures that the
feedback measurement is a representation of true force.
Noise in the sensor is also very detrimental to operation.
We use a linear potentiometer to measure spring
deflection.

\subsection{Transmission: Ball$-$Screw Selection} \label{ballScrew}
During operation, the servomotor directly drives the
ball screw, converting rotary motion to linear motion
of the ball nut. When the motor rotates, the ball nut
moves up or down the screw depending on the
direction of motor rotation. The ball nut 
pushes on the springs, which help with shock absorption and allow fine force control.

When choosing a ball screw the most important specifications are the lead and the diameter. The lead determines how fast the prosthetic will be able to retract and the diameter is proportional to the load it will be able to support.

We selected a ball screw manufactured by Thomson Linear. First we chose a 5mm/rev lead because it was the most popular for robotic applications. Additionally, it allows sufficiently fast retractability. Given an effective linear travel, ELT, of 108mm (measured using SolidWorks) and a motor speed, $\omega_m$, of 79.83 rev/sec we can calculate that time as follows:


\begin{eqnarray}
    \text{time} = \frac{ELT}{L \times \omega_m} = \frac{108 (mm) }{5 (mm/rev) \times 79.83 (rev/s)} = 0.27 s \label{eq_time} 
\end{eqnarray}


As we can see, our prosthesis can move more than 10cm in less than 3 tenths of a second. This exceeds the requirements needed for normal walking or even running.

Once we had chosen the lead, we fixed the nut size to 24mm because this ensured it would support a 350lb load, which is good enough for our 200lb person hypothesis. Finally, we selected a 24mm dia ball screw to match the nut.

\subsection{Spring Selection}

Choosing the spring
constant, $k_s$, for the elastic element requires special attention because we must balance two competing requirements:
\begin{enumerate}
\item Large force bandwidth requires a high spring constant.
\item Minimizing nonlinear friction and impedance requires a low spring constant.
\end{enumerate}
 
It is important to note that biomimetic actuators
can trade off small motion bandwidth for good force
control. This makes springs an ideal element to include in the design \cite{Robinson99LinearBiomimeticActuatorMIT}.

The following guidelines can be used when selecting a spring constant:
\begin{enumerate}
\item Define an operational bandwidth, $\omega_o$, for which the
actuator will need large forces $\rightarrow$ this places a lower bound on $k_s$
\item Insure that the controller
gains can be raised to acceptable levels of stiction
and impedance reduction $\rightarrow$ this places an upper bound on $k_s$
\end{enumerate}

Empirically, it has been shown that 315 kN/m offers an optimum balance between force bandwidth and impedance in these kind of actuators \cite{Robinson99LinearBiomimeticActuatorMIT}. Therefore, we selected this value for our springs.


\section{Socket Design: A Variable-Impedance Prosthetic Socket} \label{SocketDesign}

Surveys have shown that amputees complain about
their prosthesis being uncomfortable \cite{Ian86Socket, Nielsen90socket}. It is not
uncommon for amputees to develop skin problems on
the residual limb, such as blisters, cysts, edema, skin irritation,
and dermatitis \cite{Henke06Socket_Sores, William80Socket_skinProb}. Discomfort and skin problems are usually attributed to a poor socket fit.

The basic principles for socket design vary from
either distributing most of the load over specific loadbearing
areas or more uniformly distributing the load
over the entire limb \cite{Arthur01SocketReview}.

Even the most rigorous scientific analyses to
date have focused in large part on socket designs based on
historical use and proven clinical adequacy\cite{Arthur01SocketReview}, not on numerical analysis. Modern instrumentation
and computer modeling have allowed us to discover
what had only previously been the implied conditions
inside prosthetic sockets. However, the most recent
advances in the understanding of stresses experienced at the
limb/prosthesis interface have not yet fundamentally altered
clinical practice \cite{Sewell00Socket_Past}.

For all prosthetic socket designs, the optimal load
distribution should be proportional to the ability of the
body to sustain such stresses, without crossing the thresholds
of pain or skin breakdown \cite{Arthur01SocketReview}.

The CAD/CAM technology for the prosthetic socket
may make the socket design and manufacture process
more effective and objective. However, the current computer$-$aided
design and manufacturing
(CAD/CAM) systems cannot offer any expert suggestion
on how to make an optimal socket design.

We decided to use mirroring of compliances. This technique consists on creating a socket with varying stiffness in such a way that it mirrors the biomechanics of the underlying tissue. We chose this approach because it has been proven that an inversely proportional relationship between residual limb stiffness and the corresponding socket wall stiffness at
each spatial location across the residual limb surface leads
to reduced contact pressures 
over the fibula and tibia anatomical landmarks during walking and quiet standing \cite{HughHerr13Socket}. This is extremely important because socket
interface pressure is a major reason for sores, pain, and discomfort in sockets\cite{Faustini05Socket}.


\begin{figure}
\begin{center}
\includegraphics[width=60mm]{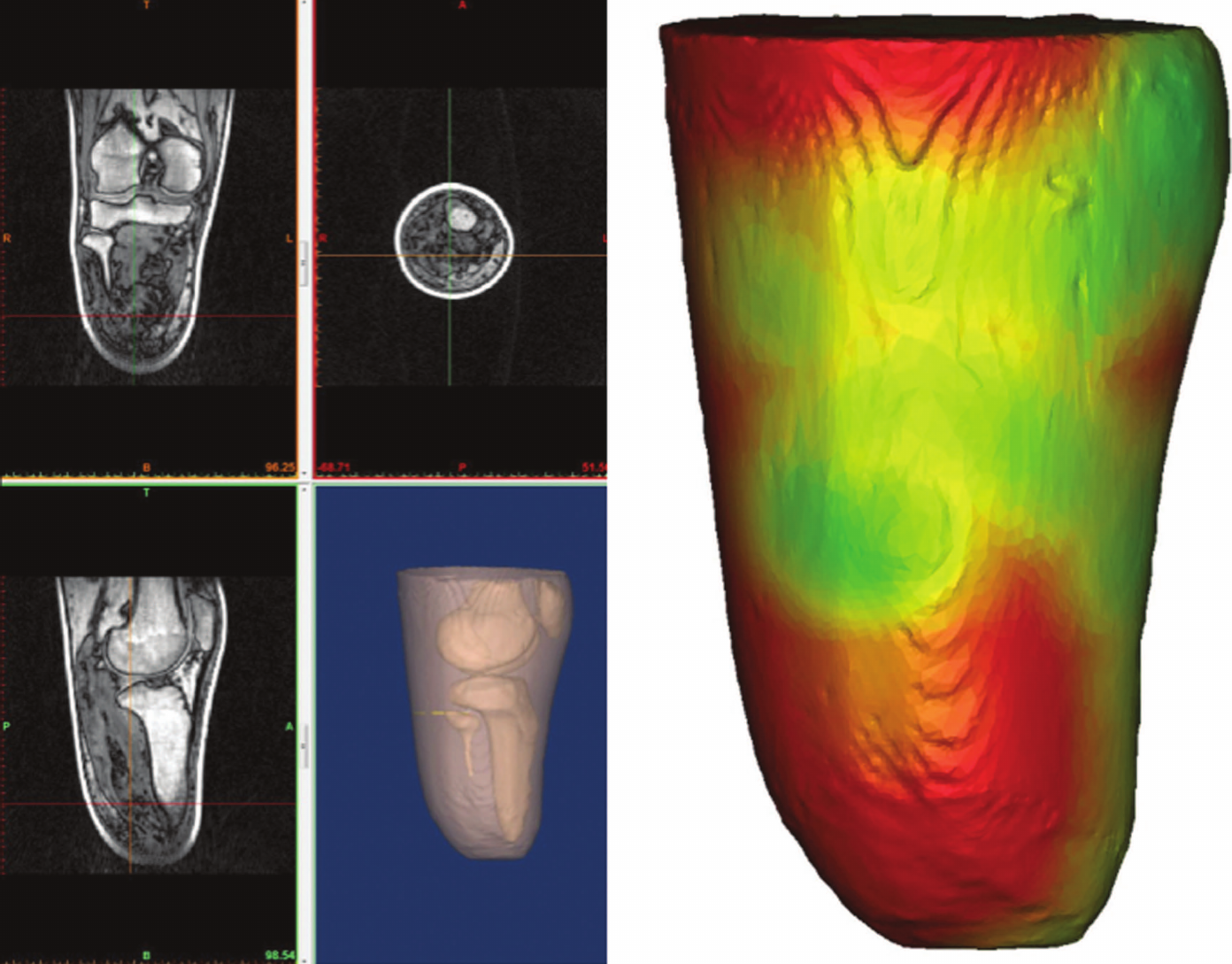}
\end{center}
\caption{Left, four MRI views of the right residual limb of the amputee participant. Upper left, anterior view; upper right, lateral view; lower left,
medial view; lower right, 3D rendering showing bones within the limb. Right, bone tissue depth representation is shown, where red denotes the
maximum bone tissue depth and green denotes the minimum depth. The bone tissue depth range is as follows: green, 0Y9mm; and red, 20Y50mm.
MRI, magnetic resonance imaging; 3D, three-dimensional. Taken from \cite{HughHerr13Socket}.\label{fig_MRI_Socket}}
\end{figure}


To create this socket there are several steps that must be followed \cite{HughHerr13Socket}:
\begin{enumerate}
\item MRI Imaging: MRI data of the amputee's
residual limb can be used to estimate
and map body stiffness and anatomical landmarks directly to
the prosthetic socket’s wall stiffness. See Fig. \ref{fig_MRI_Socket}. 

\item Inverse mapping: An inverse linear equation is used to map bone tissue depth
to socket material stiffness properties. Regions where the body
was stiffest interfaced with the most compliant material, whereas
regions where the body was softest interfaced with the least
compliant material. Empirical data has led to the following equation: $Y = 0.0382 \times X + 1.0882$
where Y is the Young’s Modulus of the printing material
and X is the bone tissue depth.

\item 3D printing: Polyjet Matrix 3D printing technology is used to seamlessly integrate
variable durometer materials into the socket design to
achieve intrinsic spatial variations in socket wall impedance
while maintaining structural integrity.
\end{enumerate}


\section{Materials}
Material selection had both cost and function in mind during the design process. Most components were either found commercially or easily manufactured with tradional manufacturing techniques. 
\begin{itemize}
\item Aluminum 6061: This material was used for parts that need to be custom machined. Aluminum 6061 provides strong structural properties, low cost, and easy machinability. Listed below are the parts that use Aluminum 6061 which can be shown in Fig. \ref{fig_explosionNAmes}:
	\begin{itemize}
	\item Motor Mounting Plate
    \item Carriage Plates
    \item Base Plate
    \item Foot Attachment
	\end{itemize}
\item Stainless Steel (Variety of Grades): This material is found in commercially found parts. The grade of the stainless steel varies depending on the product. By using commercially available parts, costs are generally lowered. The parts made from stainless steel include which can be seen in Fig. \ref{fig_explosionNAmes}:
	\begin{itemize}
	\item Screw
    \item Linear Rails
	\end{itemize}
\item ABS Plastic: ABS is a very common grade of plastic. It is heavily used in the 3D printing industry. ABS provides a low cost and durable option to custom, intricate parts. The parts using ABS plastic can be seen in Fig. \ref{fig_explosionNAmes}:
	\begin{itemize}
	\item Socket
    \item Foot
	\end{itemize}
Since the socket is to be of variable impedance, 3D printing will be utilized. In addition, the foot can be readily 3D printed with different designs to provide the ultimate level of comfort for the user. Once determined which is the best model, the foot can be easily produced using an injection molding manufacturing process to reduce cost if scale up occurs. 
\item Rubber: A variant of rubber will be used for the sole of the foot. This will allow traction to be generated similar to a sneaker.
\item Teflon: Teflon coated linear bearings are to be used as a cheap, maintenance free alternative to conventional linear bearings.
    
\end{itemize}

\section{Finite Element Analysis}
\begin{figure}
\begin{center}
\includegraphics[width=60mm]{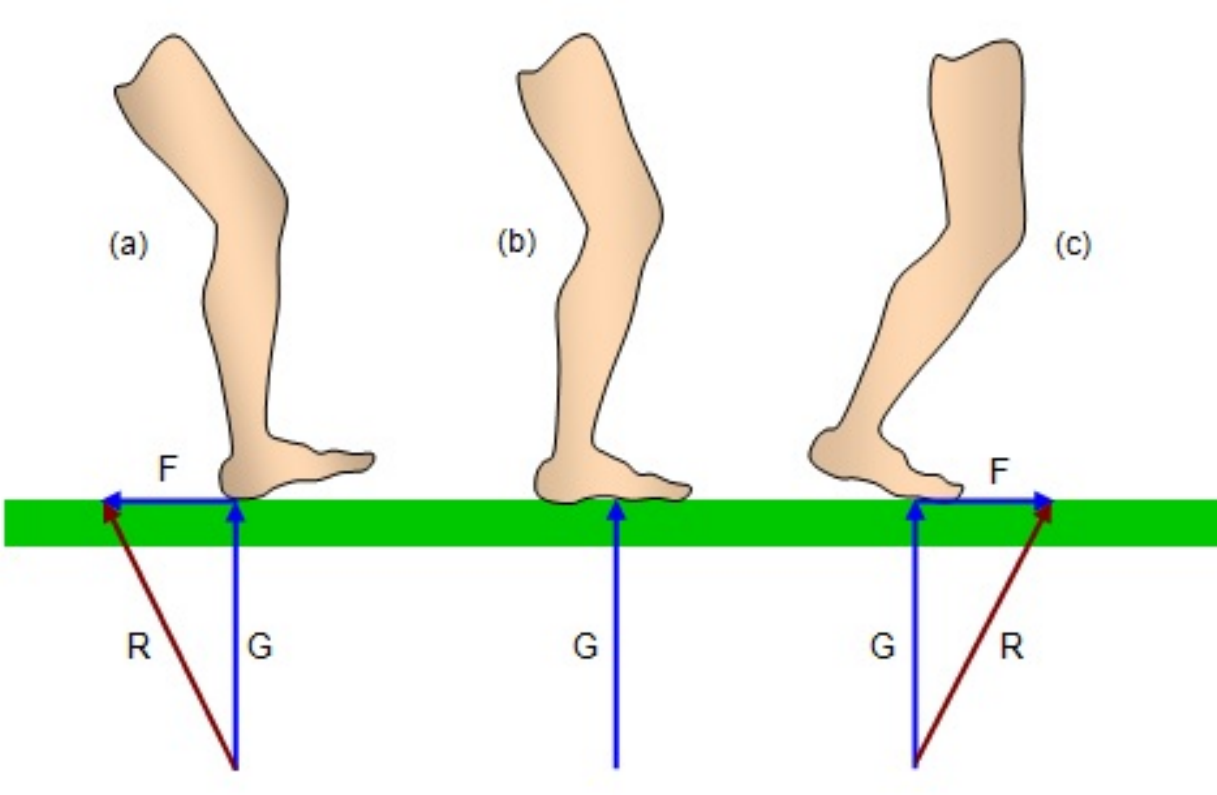}
\end{center}
\caption{Simplified force model used during the finite element analysis. Taken from \cite{ForceWalking}}
\label{fig_force}
\end{figure}

After several iterations of the design were completed, a finite elemental analysis was used to determine the structural integrity of the robot.  Both the magnitude and direction of walking forces were approximated during the simulation. Fig. \ref{fig_force} shows the forces to be simulated throughout a typical walk cycle. Each of the forces (or sum of forces) were simulated at 1.5x the arbitrary body weight of 200 lbs.

several assumptions, which aren’t necessarily accurate, were made during the FE simulations. These assumptions include a global bond between all connections, essentially making the whole robot a rigid body.

Analyzing the simulations, the current design has the potential to fail depending on the validity of the assumptions. Testing would ultimately have to be conducted to ensure that the model presented is safe. Assuming structural failure, the highest concentrations would be seen at the linear rails between the base plate and foot attachment. This makes logical sense as this portion of the design is under intensive moment loads. Possibly remedies include geometry changes, material changes, architecture changes, etc.

\begin{figure}
\begin{center}
\includegraphics[width=60mm]{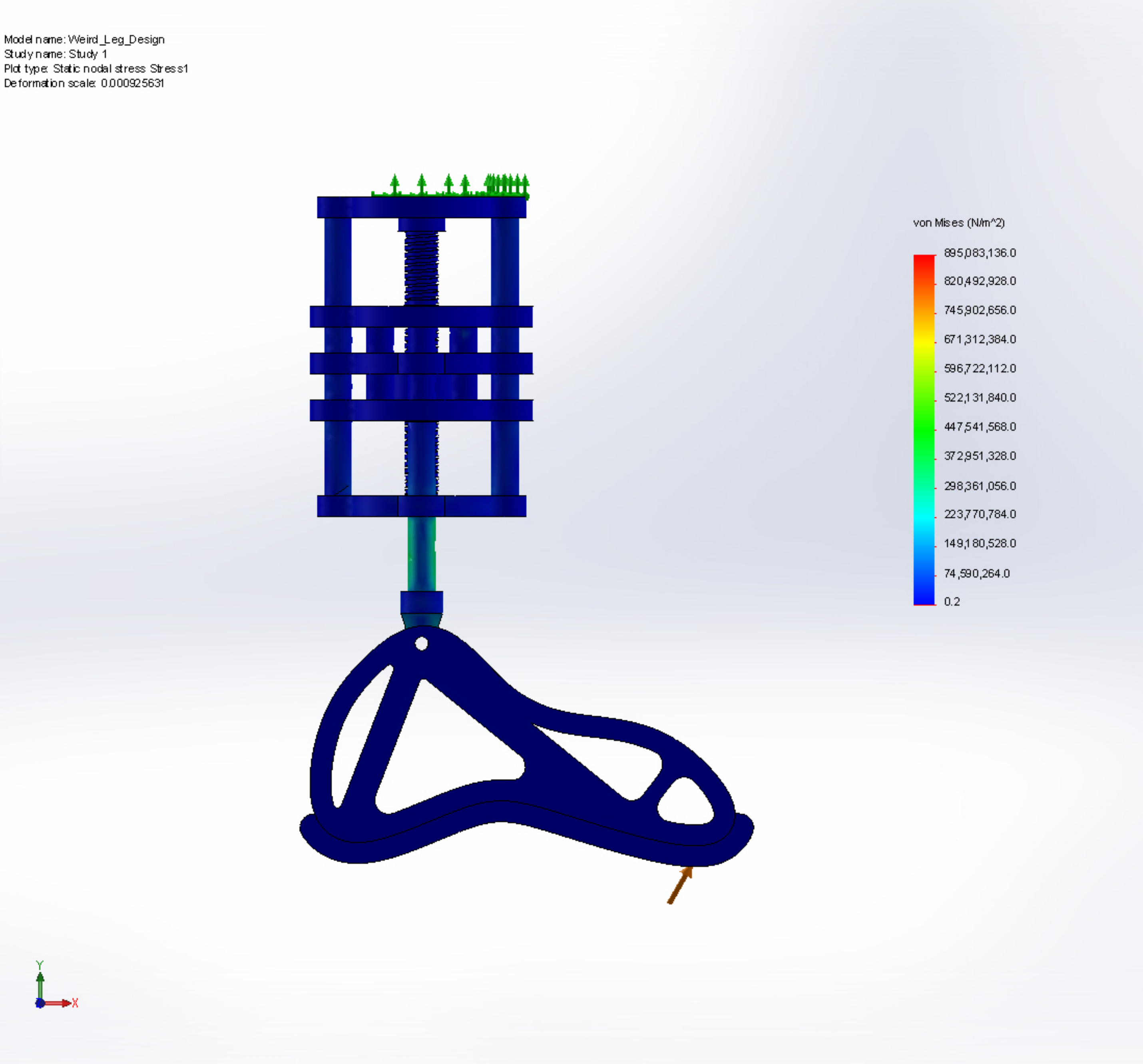}
\end{center}
\caption{Finite element analysis of stress distribution during opposite heel strike. See Fig.\ref{fig_gaitPaper} for walking cycle notation.}
\label{fig_FE_ball}
\end{figure}

\begin{figure}
\begin{center}
\includegraphics[width=60mm]{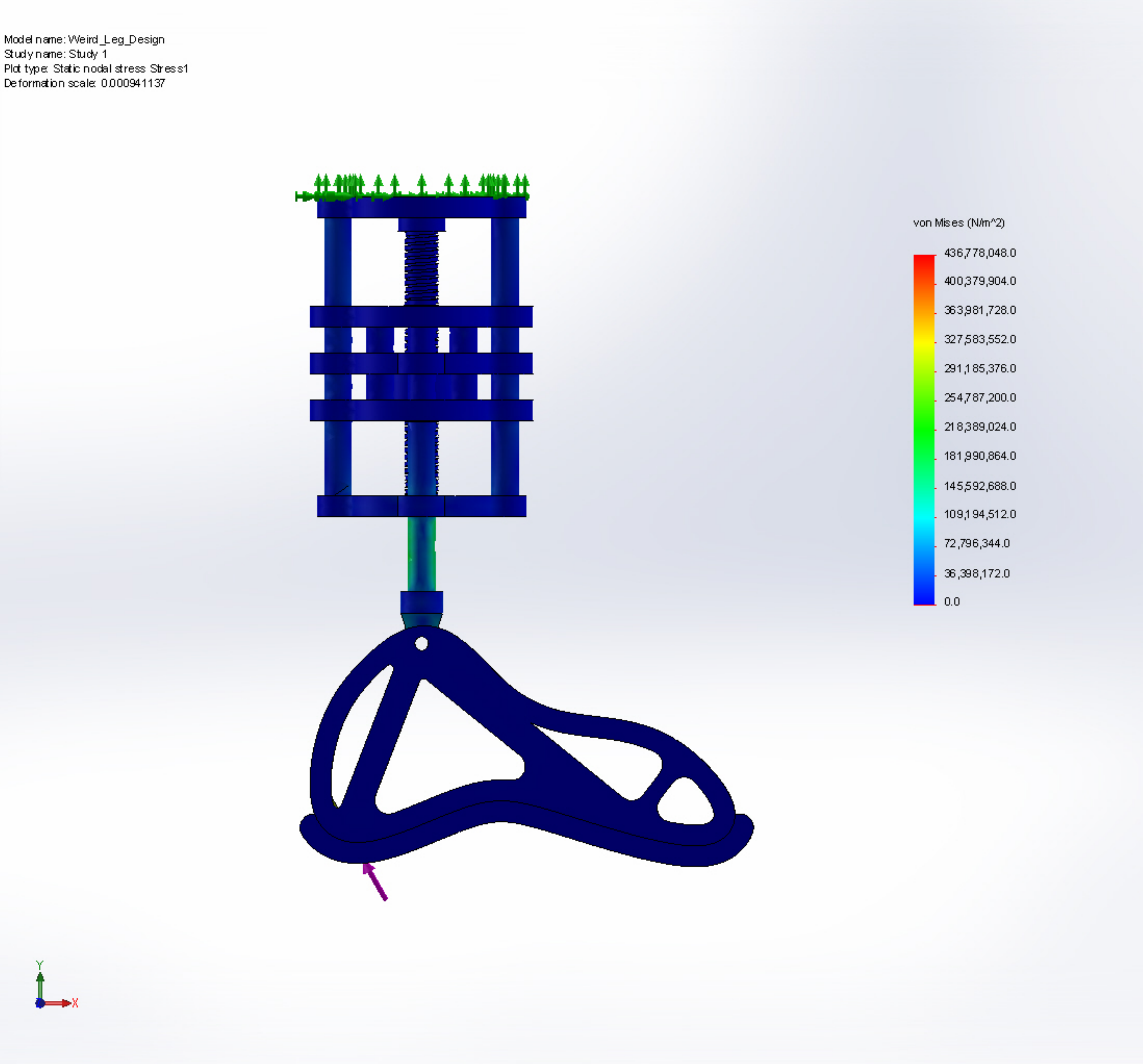}
\end{center}
\caption{Finite element analysis of stress distribution during heel strike. See Fig.\ref{fig_gaitPaper} for walking cycle notation.}
\label{fig_FE_heel}
\end{figure}

\begin{figure}
\begin{center}
\includegraphics[width=60mm]{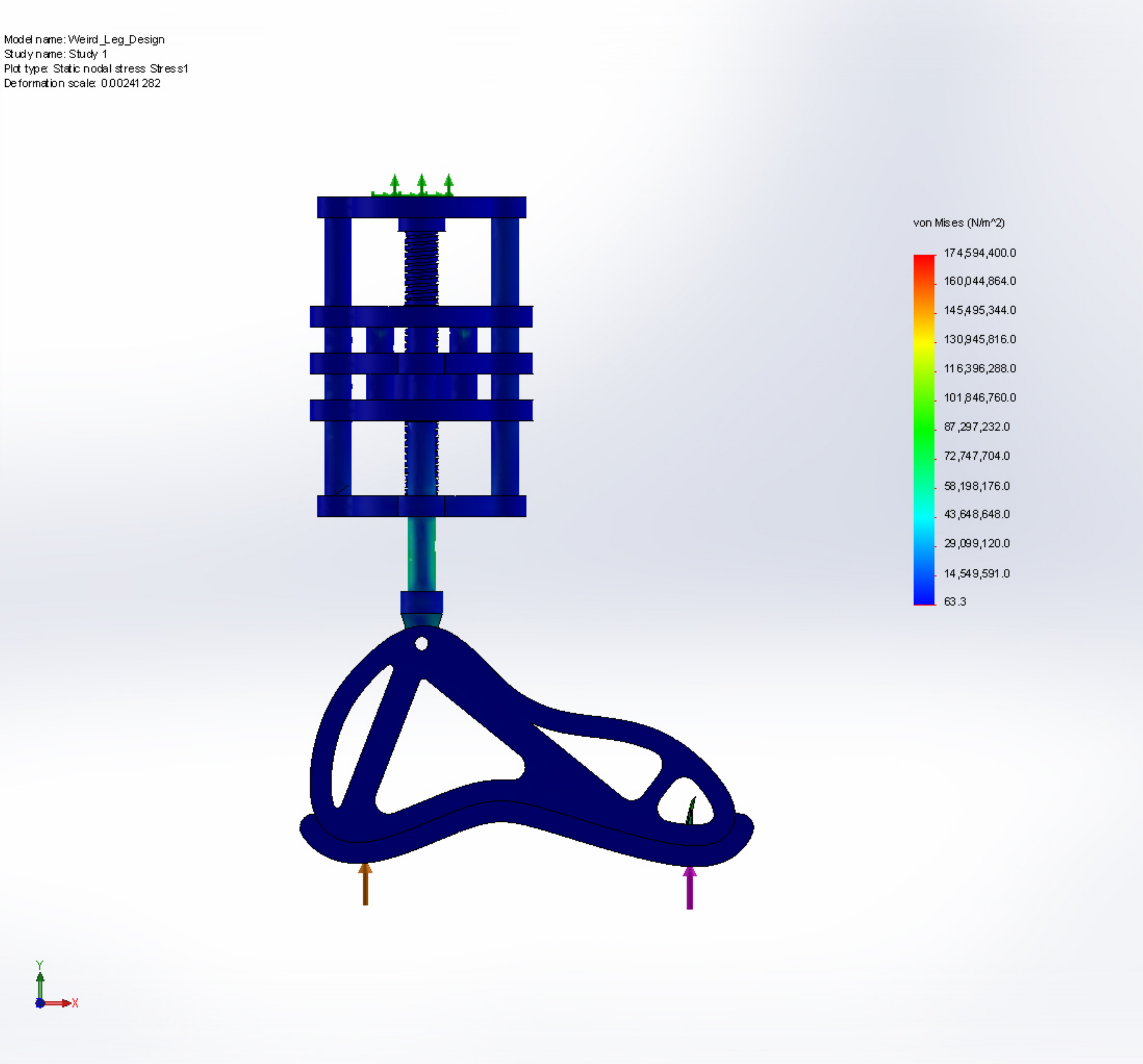}
\end{center}
\caption{Finite element analysis of stress distribution while standing for a 200lb person.}
\label{fig_FE_stand}
\end{figure}

Fig. \ref{fig_FE_ball}, Fig. \ref{fig_FE_heel} and Fig. \ref{fig_FE_stand} show the results of the simulation. The color bar shows the Von Mises stress, which is used as an indicator to know to predict yielding of materials. A material is said to start yielding when its von Mises stress (which can be computed using the Cauchy stress tensor) reaches a critical value known as the yield strength. At that precise point the material will begin to deform plastically (yield), that is, it will undergo non$-$reversible changes of shape due to the applied forces.

\section{Conclusions}
We presented a prosthesis for transfemoral amputees whose general specifications can be found in  Table. \ref{table_generalSpecs}

Overall, the use of a series elastic actuator coupled with force control provides a potential for a low cost, natural option to transfemoral amputees. With further testing, this design could show promise amongst less financially fortunate patients. Utilizing cheap sensors, cheap materials, and commercialized products the financial burden of purchasing an assistive robot may be overcome.

Analyzing the specs in Table. \ref{table_generalSpecs}, the design shows promise to be a comfortable, natural, and versatile alternative to a conventional $“$peg-leg$”$.



 	\begin{table}[]
 		\begin{center}
 		\caption{Prosthesis Specifications}
 		\label{table_generalSpecs}
 		\begin{tabular}{|c|c|}
 			\hline
 			Socket Design & Variable impedance  \\
 			\hline
 			Foot Design & User Specific or Mass Produced\\
 			\hline
			Overall Weight  & 9 Kg\\
 			\hline
 			Overall Length (resting position) & 665 mm\\
 			\hline
            Effective Travel & 108 mm\\
 			\hline
            Sensors & Linear Potentiometers\\
 			\hline
 		\end{tabular}
 		\end{center}
 	\end{table}

\bibliographystyle{IEEEtran}	
\bibliography{RetractableProsthRefs}

\begin{thebibliography}{10}
\providecommand{\url}[1]{#1}
\csname url@samestyle\endcsname
\providecommand{\newblock}{\relax}
\providecommand{\bibinfo}[2]{#2}
\providecommand{\BIBentrySTDinterwordspacing}{\spaceskip=0pt\relax}
\providecommand{\BIBentryALTinterwordstretchfactor}{4}
\providecommand{\BIBentryALTinterwordspacing}{\spaceskip=\fontdimen2\font plus
\BIBentryALTinterwordstretchfactor\fontdimen3\font minus
  \fontdimen4\font\relax}
\providecommand{\BIBforeignlanguage}[2]{{%
\expandafter\ifx\csname l@#1\endcsname\relax
\typeout{** WARNING: IEEEtran.bst: No hyphenation pattern has been}%
\typeout{** loaded for the language `#1'. Using the pattern for}%
\typeout{** the default language instead.}%
\else
\language=\csname l@#1\endcsname
\fi
#2}}
\providecommand{\BIBdecl}{\relax}
\BIBdecl

\bibitem{GauthierGagnon99UseOfTransfemoralProsth}
\BIBentryALTinterwordspacing
C.~Gauthier-Gagnon, M.-C. Grise, and D.~Potvin, ``Enabling factors related to
  prosthetic use by people with transtibial and transfemoral amputation,''
  \emph{Archives of Physical Medicine and Rehabilitation}, vol.~80, no.~6, pp.
  706 -- 713, 1999. [Online]. Available:
  \url{http://www.sciencedirect.com/science/article/pii/S0003999399901776}
\BIBentrySTDinterwordspacing

\bibitem{Pearlman08Cheap}
J.~Pearlman, R.~Cooper, M.~Krizack, A.~Lindsley, Y.~Wu, K.~Reisinger,
  W.~Armstrong, H.~Casanova, H.~Chhabra, and J.~Noon, ``Lower-limb prostheses
  and wheelchairs in low-income countries [an overview],'' \emph{Engineering in
  Medicine and Biology Magazine, IEEE}, vol.~27, no.~2, pp. 12--22, March 2008.

\bibitem{Pratt95SeriesElasticAct}
G.~Pratt and M.~Williamson, ``Series elastic actuators,'' in \emph{Intelligent
  Robots and Systems 95. 'Human Robot Interaction and Cooperative Robots',
  Proceedings. 1995 IEEE/RSJ International Conference on}, vol.~1, Aug 1995,
  pp. 399--406 vol.1.

\bibitem{Matthew95SerElastActThesis}
M.~M. Williamson, ``Series elastic actuators,'' Master's thesis, Massachusetts
  Institute of Technology, 1995.

\bibitem{Asada87DirectDriveBOOK}
H.~Asada and K.~Youcef-Toumi, \emph{Direct Drive Robots: Theory and
  Practice}.\hskip 1em plus 0.5em minus 0.4em\relax MIT Press, 1987.

\bibitem{pratt97SpringTurkey}
J.~Pratt, P.~Dilworth, and G.~Pratt, ``Virtual model control of a bipedal
  walking robot,'' in \emph{Robotics and Automation, 1997. Proceedings., 1997
  IEEE International Conference on}, vol.~1, Apr 1997, pp. 193--198 vol.1.

\bibitem{pratt98SpringFlamingo}
J.~Pratt and G.~Pratt, ``Intuitive control of a planar bipedal walking robot,''
  in \emph{Robotics and Automation, 1998. Proceedings. 1998 IEEE International
  Conference on}, vol.~3, May 1998, pp. 2014--2021 vol.3.

\bibitem{BD08BigDog}
M.~Raibert, K.~Blankespoor, G.~Nelson, R.~Playter, and the BigDog~Team,
  ``Bigdog, the rough-terrain quadruped robot,'' in \emph{Proceedings of the
  17th World Congress The International Federation of Automatic Control}, 2008.

\bibitem{Arumugom09ModelingLinearActutator}
Arumugom.S, Muthuraman.S, and Ponselvan.V, ``Modeling and application of series
  elastic actuators for force control multi legged robots,'' \emph{Journal of
  Computing}, vol.~1, pp. 26--33, 2009.

\bibitem{Arthur01SocketReview}
A.~F. Mak, M.~Zhang, and D.~A. Boone, ``State-of-the-art research in lower-limb
  prosthetic biomechanicssocket interface: A review,'' \emph{Journal of
  Rehabilitation Research and Development}, vol.~38, pp. 161--174, 2001.

\bibitem{Radcliff61Patellar_Socket}
R.~CW and F.~J., ``The patellar-tendon-bearing below-knee prosthesis,''
  Berkeley, CA: Biomechanics laboratory, University of California, Tech. Rep.,
  1961.

\bibitem{Radcliff55Fitting_Socket}
R.~CW., ``Functional considerations in the fitting of aboveknee prostheses,''
  \emph{Artificial Limb}, vol.~2, pp. 35--60, 1955.

\bibitem{Fillauer89Socket}
F.~CE, P.~CH, and F.~KD., ``Evaluation and development of the silicone suction
  socket (3s) for below-knee prostheses.'' \emph{The Journal of the
  International Society for Prosthetics and Orthotics}, vol.~22, pp. 92--103,
  1989.

\bibitem{Kristinsson93Socket}
K.~O., ``The iceross concept: a discussion of a philosophy,'' \emph{Prosthetics
  and Orthotics International}, vol.~17, pp. 49--55, 1993.

\bibitem{Dollar008LowerExtremStateofArt}
A.~Dollar and H.~Herr, ``Lower extremity exoskeletons and active orthoses:
  Challenges and state-of-the-art,'' \emph{Robotics, IEEE Transactions on},
  vol.~24, no.~1, pp. 144--158, Feb 2008.

\bibitem{ForceWalking}
K.~Gibbs. (2013) The forces on the body during walking and running.

\bibitem{BallScrewEfficiency}
\BIBentryALTinterwordspacing
M.~Budimir. Video. Design World. [Online]. Available:
  \url{http://www.youtube.com/watch?v=XSbXRIigA0g}
\BIBentrySTDinterwordspacing

\bibitem{Robinson99LinearBiomimeticActuatorMIT}
D.~Robinson, J.~Pratt, D.~Paluska, and G.~Pratt, ``Series elastic actuator
  development for a biomimetic walking robot,'' in \emph{Advanced Intelligent
  Mechatronics, 1999. Proceedings. 1999 IEEE/ASME International Conference on},
  1999, pp. 561--568.

\bibitem{eppinger87BandwidthLimit}
S.~Eppinger and W.~Seering, ``Understanding bandwidth limitations in robot
  force control,'' in \emph{IEEE International Conference on Robotics and
  Automation}, 1987.

\bibitem{Ian86Socket}
I.~McColl, \emph{Review of Artificial Limb and Appliance Centre Services: The
  Report of an Independent Working Party Under the Chairmanship of
  Professor}.\hskip 1em plus 0.5em minus 0.4em\relax Great Britain. Department
  of Health and Social Security, 1986, vol.~2.

\bibitem{Nielsen90socket}
N.~CC., ``A survey of amputees: functional level and life satisfaction,
  information needs, and the prosthetist?s role.'' \emph{Journal of Prosthetics
  and Orthotics}, vol.~3, pp. 125--9, 1990.

\bibitem{Henke06Socket_Sores}
H.~E.~J. MEULENBELT, P.~U. DIJKSTRA, M.~F. JONKMAN4, and J.~H.~B. GEERTZEN,
  ``Skin problems in lower limb amputees: A systematic review,''
  \emph{Disability and Rehabilitation}, vol.~28, pp. 603--608, 2006.

\bibitem{William80Socket_skinProb}
S.~W. Levy, ``Skin problems of the leg amputee,'' \emph{Prosthetics and
  Orthotics International}, vol.~4, pp. 37 -- 44, 1980.

\bibitem{Sewell00Socket_Past}
S.~P, N.~S, V.~J, and A.~S., ``Development in the transtibial prosthetic socket
  fitting process: a review of past and present research.'' \emph{Prosthetics
  and Orthotics International}, vol.~24, pp. 97--107, 2000.

\bibitem{HughHerr13Socket}
D.~M. Sengeh and H.~Herr, ``A variable-impedance prosthetic socket for a
  transtibial amputee designed from magnetic resonance imaging data,''
  \emph{Journal of Prosthetics and Orthotics}, vol.~25, pp. 129--137, 2013.

\bibitem{Faustini05Socket}
F.~MC, C.~RH, N.~RR, and et~al., ``Design and analysis of orthogonally
  compliant features for local contact pressure relief in transtibial
  prostheses.'' \emph{Journal of Biomechanical Engineering}, vol. 127, pp.
  946--951, 2005.

\end{thebibliography}

\end{document}